\documentclass[journal]{IEEEtran}

\usepackage[noadjust]{cite}
\usepackage{graphicx}
\usepackage{amsmath}
\usepackage{amssymb}
\usepackage{multirow}
\usepackage[letterpaper=true,colorlinks,bookmarks=false]{hyperref}
\usepackage{subfigure}
\usepackage[lined,boxed, ruled]{algorithm2e}

\begin{document}

\title{Disc-aware Ensemble Network for Glaucoma Screening from Fundus Image}

\author{Huazhu Fu, Jun Cheng, Yanwu Xu, Changqing Zhang,  Damon Wing Kee Wong, Jiang Liu, and Xiaochun Cao
	\thanks{H.~Fu and D.~W.~K.~Wong are with Institute for Infocomm Research, Agency for Science, Technology and Research, Singapore 138632 (e-mail: huazhufu@gmail.com, wkwong@i2r.a-star.edu.sg).}
	\thanks{J.~Cheng and J.~Liu are with the Cixi Institute of Biomedical Engineering, Chinese Academy of Sciences, Zhejiang 315201, China (e-mail: sam.j.cheng@gmail.com, jimmyliu@nimte.ac.cn)}
	\thanks{Y.~Xu is with CVTE Research, Guangzhou Shiyuan Electronic Technology Company Limited, Guangzhou 510530, China (e-mail: xuyanwu@cvte.com).}
	\thanks{C.~Zhang is with the School of Computer Science and Technology, Tianjin University, Tianjin 300072, China (e-mail: zhangchangqing@tju.edu.cn)}
	\thanks{X.~Cao is with the State Key Laboratory of Information Security, Institute of Information Engineering, Chinese Academy of Sciences, Beijing 100093, China (e-mail: caoxiaochun@iie.ac.cn)}}

\maketitle

\begin{abstract}
	
Glaucoma is a chronic eye disease that leads to irreversible vision loss. Most of the existing automatic screening methods firstly segment the main structure, and subsequently calculate the clinical measurement for detection and screening of glaucoma. However, these measurement-based methods rely heavily on the segmentation accuracy, and ignore various visual features. In this paper, we introduce a deep learning technique to gain additional image-relevant information, and screen glaucoma from the fundus image directly. Specifically, a novel Disc-aware Ensemble Network (DENet) for automatic glaucoma screening is proposed, which integrates the deep hierarchical context of the global fundus image and the local optic disc region. Four deep streams on different levels and modules are respectively considered as global image stream, segmentation-guided network, local disc region stream, and disc polar transformation stream. Finally, the output probabilities of different streams are fused as the final screening result. The experiments on two glaucoma datasets (SCES and new SINDI datasets) show our method outperforms other state-of-the-art algorithms.
	
\end{abstract}

\begin{IEEEkeywords}
	Deep learning, glaucoma screening, optic disc segmentation, neural network.
\end{IEEEkeywords}

\IEEEpeerreviewmaketitle

\section{Introduction}
 
\begin{figure}[!t]
	\begin{center}
		\includegraphics[width=1\linewidth]{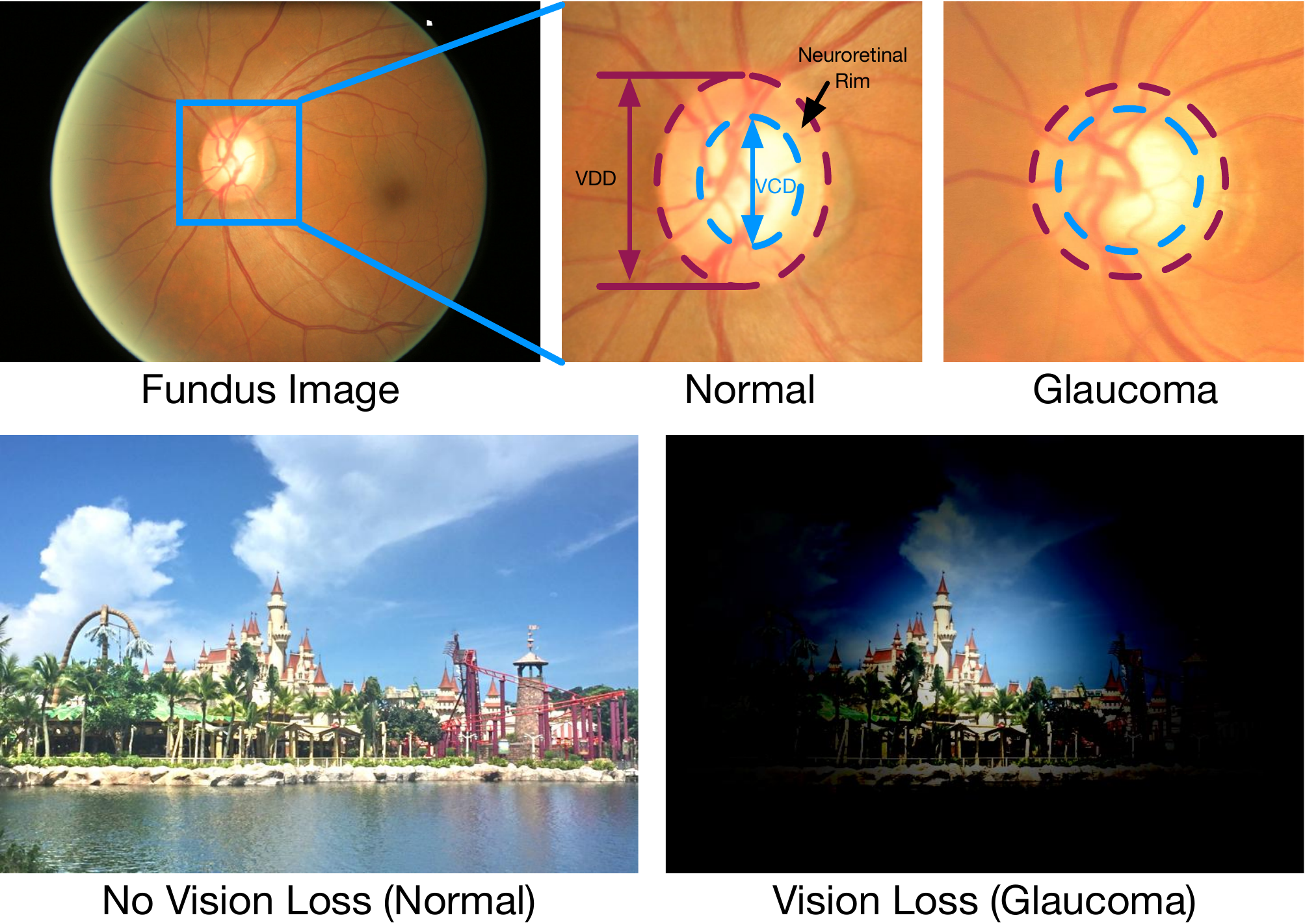}
		\caption{Top: the whole fundus image and zoom-in normal/glaucoma disc regions, where the vertical cup to disc ratio (CDR) is calculated by the ratio of vertical cup diameter (VCD) to vertical disc diameter (VDD). Bottom: the visual fields with normal and glaucoma cases.}
		\label{img-cover}
	\end{center}
\end{figure}

Glaucoma is one of the major leading causes of blindness among eye diseases, predicted to affect around 80 million people by 2020~\cite{Tham2014}. Unlike other eye diseases such as cataracts and myopia, vision loss from glaucoma cannot be reversed. Early screening is thus essential for early treatment to preserve vision and maintain life quality. However,  many glaucoma patients are not aware of their condition~\cite{Shen2008IOVS}. That is why glaucoma is also called the ``silent theft of sight'', as shown in the bottom row of Fig.~\ref{img-cover}. 
Clinically, there are three examinations practiced to screen glaucoma: intraocular pressure (IOP) measurement, function-based visual field test, and optic nerve head (ONH) assessment. IOP is an important risk factor but not specific enough to be an effective detection tool for a great number of glaucoma patients with normal tension. Function-based visual field testing requires specialized perimetric equipment not normally present in primary healthcare clinics. Moreover, the early glaucoma often does not have visual symptoms. ONH assessment is a convenient way to detect glaucoma early, and is currently performed widely by trained glaucoma specialists~\cite{Jonas1999,Morgan879,Fu2017TMI}. 

Manual ONH assessment by trained clinicians is time-consuming and costly. Thus, an automatic  method is necessity for screening. One popular ONH assessment method is based on the  measurement of clinical parameters, such as the vertical cup to disc ratio (CDR)~\cite{Jonas2000}, rim to disc area ratio, and disc diameter~\cite{HANCOXOD199959}.  Among them, CDR is well accepted and commonly used by clinicians.   As shown in the top row of Fig.~\ref{img-cover}, the CDR is calculated by the ratio of vertical cup diameter (VCD) to vertical disc diameter (VDD). In general, a larger CDR suggests a higher risk of glaucoma and vice versa. For automatic screening, measurement-based methods have been proposed~\cite{Joshi2011,Yin2011,Cheng2013,Cheng2017SVM,Fu2018TMI}, which segment the main structure (e.g., optic disc and optic cup) first, and then calculate the clinical measurement values to identify the glaucoma cases. For example, a superpixel-based classifier with various handcrafted visual features is utilized to extract the optic disc and cup regions~\cite{Cheng2013}. The CDR value is then calculated based on the segmented regions. In~\cite{Cheng2015TBME}, a CDR assessment using fundus image is proposed, where a sparse dissimilarity-constrained coding approach is employed to consider both the dissimilarity constraint and the sparsity constraint from a set of reference discs with known CDRs. The reconstruction coefficients are used to compute the CDR for the testing disc. Recently, the deep learning based techniques have been also introduced to segment the fundus image. 
A multi-label deep network is proposed in~\cite{Fu2018TMI} to segment the optical disc and cup jointly. Then the CDR is accordingly calculated to verify the glaucoma detection.  However, these measurement-based methods rely heavily on the segmentation accuracy, which are easily affected by pathological regions and low contrast quality. Moreover, some clinical measurements identify a wider set of image properties, some of which are unrelated to what clinicians seem to recognize as relevant.

In contrast,  learning-based methods have been recently  proposed to utilize the various visual features from the fundus image  and screen the glaucoma disease directly through a learned classifier~\cite{Bock2010MIA,MOOKIAH201273,Noronha2014,Acharya2015}. The extracted visual features could explore more image relevant information, and have more representation capacity than clinical measurements. Noronha \textit{et al.}~\cite{Noronha2014}  proposed an automated glaucoma diagnosis method using higher order spectra  cumulants extracted from Radon transform applied on digital fundus images. Besides, Acharya \textit{et al.}~\cite{Acharya2015}  proposed a screening method using various features extracted from Gabor transform applied on digital fundus images.
Deep learning techniques have been recently demonstrated to yield highly discriminative representations that have aided  a considerable number of  computer vision applications~\cite{Krizhevsky2012,LeCun2015} and medical image analysis areas~\cite{Gulshan2016,Esteva2017,ShuWeiTing2017}. For example, Convolutional Neural Networks (CNNs) learn hierarchical features from raw image data and are demonstrated to achieve a high performance on retinal vessel detection~\cite{Li2015TMI,Fu2016MICCAI}. For glaucoma detection, works in~\cite{Chen2015MICCAI,Orlando2016} applied deep learning systems to detect glaucoma from fundus images directly. However, these methods merely focus on optic disc region limited contextual information contained in the patch, and thus ignore the global information from the whole fundus image. The accuracy of optic disc region cropping disturbs the detection performance. 

In this paper,  we propose a glaucoma screening network for the fundus image with the following aspects:
\begin{enumerate}
	\item \textbf{Disc-aware:} Glaucoma is an eye disease that is often associated with elevated intraocular pressure, which damages the optic nerve (disc). Thus the main clinical symptoms appear in the optic disc region~\cite{Garway-Heath352,LESKE200885}. Based on this, the proposed network should highlight the contextual information in the optic disc region.
	\item \textbf{Multi-level:} Most of existing deep learning networks focus on the global image directly, and employ pooling the layer to achieve hierarchical representations~\cite{Krizhevsky2012,Gulshan2016}. A common limitation of deep learning based approaches is the need to downsample the input image to a low resolution (i.e.,  224$\times$224) in order for the network size to be computationally manageable. However, this downsampling leads to loss of image details that are important for discrimination of subtle pathological changes. In comparison,  the local region scale preserves a fine representation, but ignores the global structure information. Thus incorporating both local and global contexts could effectively improve the performance.
	\item \textbf{Multi-module:} In clinical trials, fine measurements are often used to evaluate disease indicator. The geometric transformation could alter the morphology of the pathological region and enhance the detection changing~\cite{Salehinejad2018}. A good example is the optic disc and cup segmentation~\cite{Fu2018TMI}, in which polar transformation is employed to balance the region proportion and improve the optic cup segmentation performance. In our work, what we consider is  whether the geometry transformation could improve the screening performance.  
\end{enumerate}  

\begin{figure*} [!t]
	\begin{center}
		\includegraphics[width=0.95\linewidth]{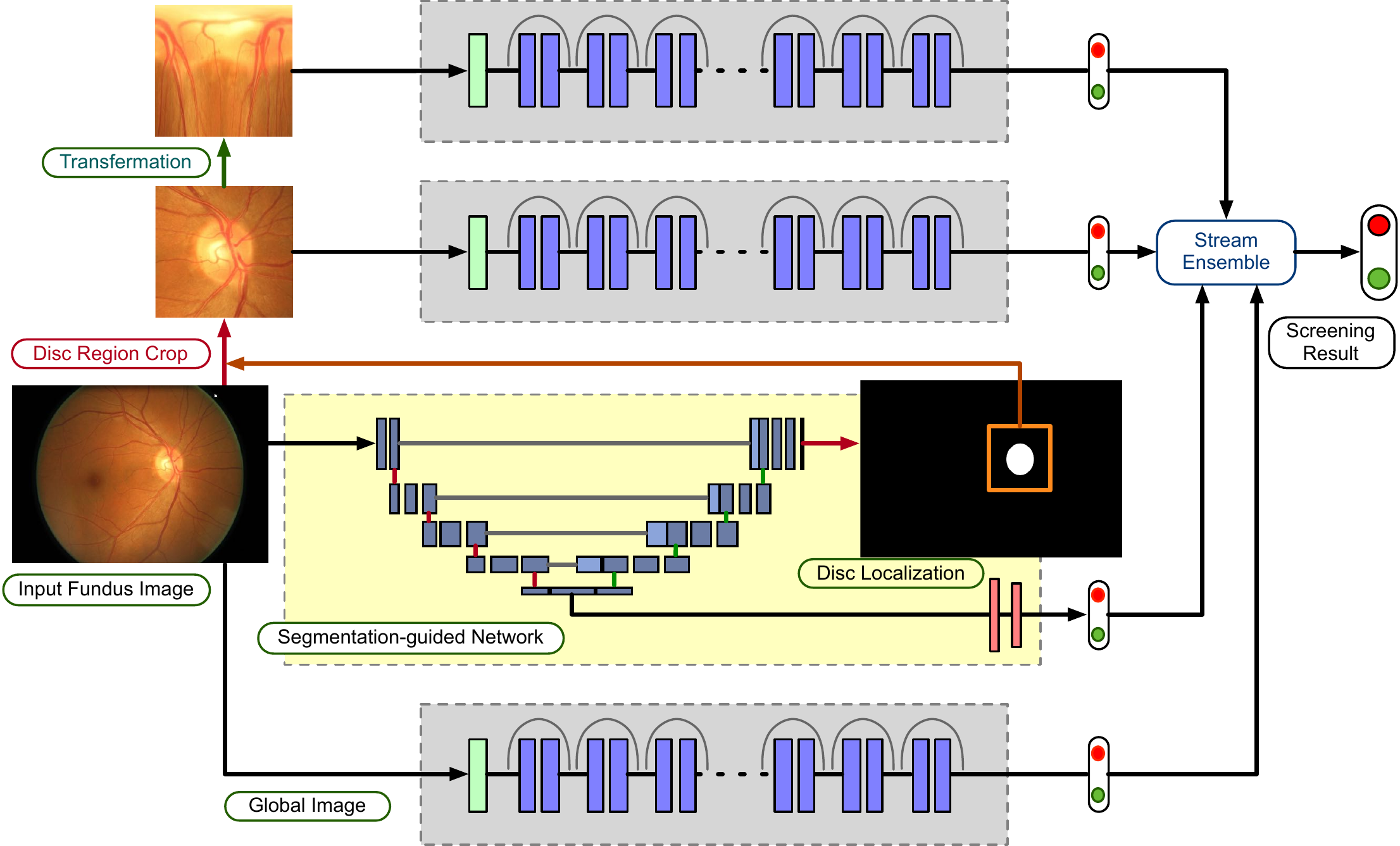}
		\caption{Architecture of our DENet, which contains four streams: global image stream produces the result based on the global fundus image; the segmentation-guided network localizes the optic disc region and generates a detection output embedded the disc-segmentation representation; disc region stream works on disc region cropped by disc segmentation map from segmentation-guided network; disc polar stream transfers the disc region image into the polar coordinate system. The combination of these four streams is fused as the final glaucoma screening result.}
		\label{img-framework}
	\end{center}
\end{figure*}

To address these, we propose a novel Disc-aware Ensemble Network (DENet) for automatic  glaucoma screening, which contains four deep streams corresponding to various levels and modules of the fundus image, as shown in Fig.~\ref{img-framework}. The first one is a global image stream, which represents the global fundus structure on image level, and works as a classification on the fundus image directly. The second one is a segmentation-guided network, which detects the disc localization from the whole fundus image and embeds the disc segmentation representation to detect glaucoma at the image level. The third stream is based on the local disc region, which produces the screening probability from the disc region level. The final stream focuses on disc region with polar transformation, which enlarges the disc and cup structure with the geometry operation and improves the screening performance.  Finally, all the outputs of these deep streams are combined to obtain the final screening result. Our proposed DENet is an automatic system for the whole fundus image that includes disc detection and glaucoma screening, and uses the hierarchical representations on different levels (global image and disc region) and modules (Cartesian and Polar coordinates). For evaluation, we test our method on two clinical datasets, the SCES dataset and a new collected SINDI dataset. The experiments show that our method outperforms  state-of-the-art methods on glaucoma screening.
The main contributions of this paper are as follows: 
\begin{enumerate}
	\item An ensemble deep network is proposed for glaucoma screening by considering various levels and modules of the fundus image. Our method achieves promising results as it can make further improvement based on the existing networks. Another advantage of our ensemble method is that it can be flexibly embedded within various networks, which makes it easier to adapt to different scenarios.
	\item Beside the ensemble architecture, we also provide a novel Segmentation-guided Network, which localizes the disc region and generates the screening result by embedding the disc-segmentation representation.
	\item For evaluation of glaucoma screening, we construct a new dataset, called the Singapore Indian Eye Study (SINDI) dataset, which is a population-based study conducted and contains a total of 5783 eye images including 113 glaucomatous eyes and  5670 normal eyes. 
\end{enumerate}

\section{Disc-aware Ensemble Network}
\label{sec-method}

Our Disc-aware Ensemble Network (DENet) takes into account two levels of fundus image information, that is the global image and the local disc region, as shown in Fig.~\ref{img-framework}. The global image level provides the coarse structure representation on the whole fundus image, while the local disc region is utilized to learn a fine representation around the optic disc.

\begin{figure*}[!t]
\centering
\subfigure[Segmentation-guided Network]{ \includegraphics[width=0.95\linewidth]{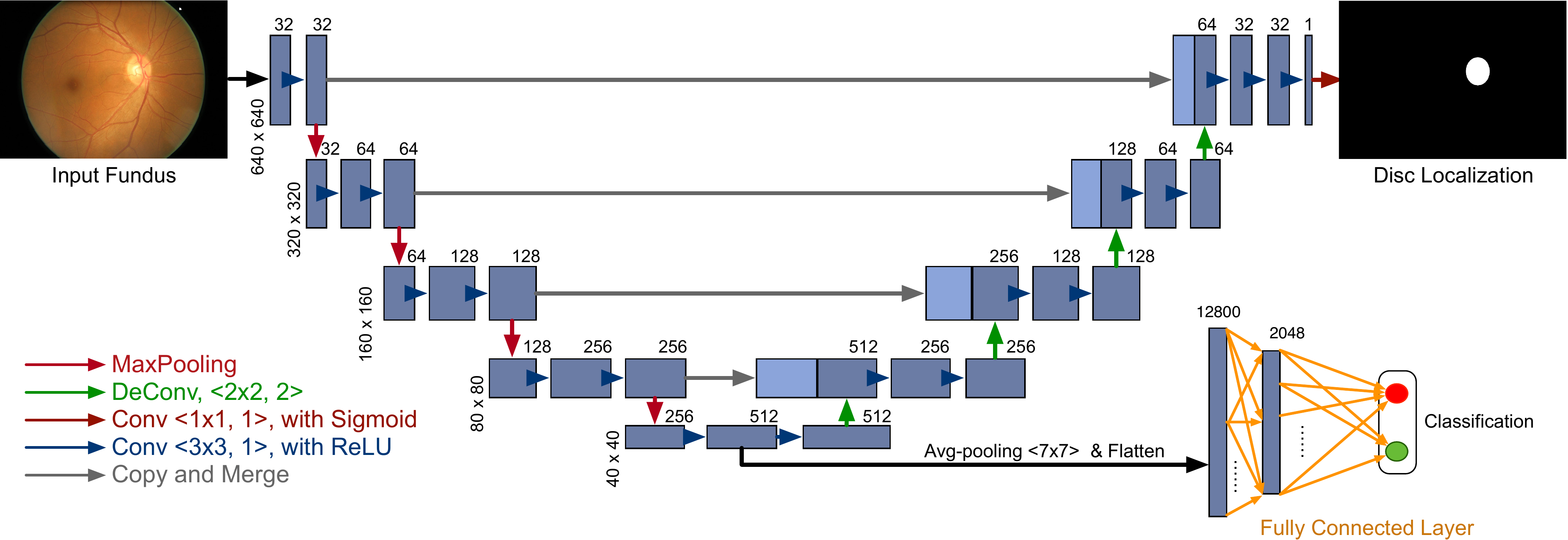}  } \\
\subfigure[ResNet-50 Network]{ \includegraphics[width=0.9\linewidth]{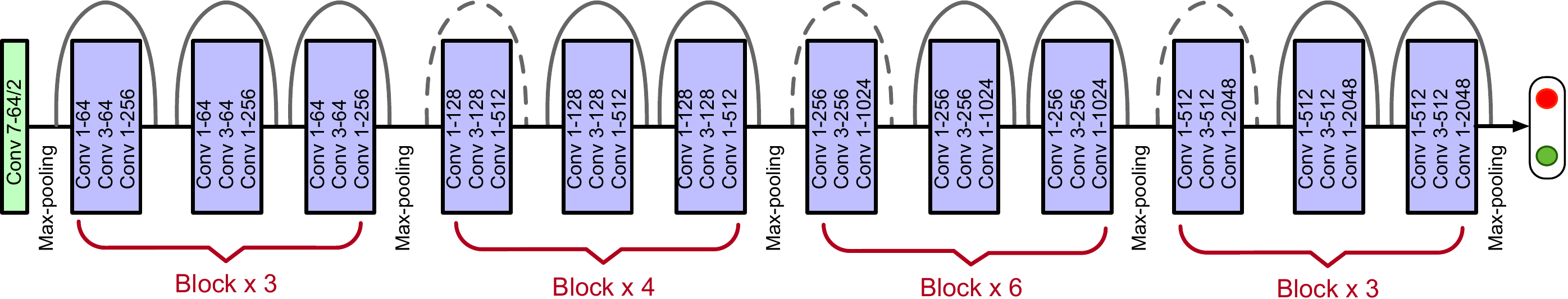} }  
\caption{(A) The detailed architecture of our segmentation-guided network, which is based on U-shape convolutional network to detect optic disc region and guide glaucoma screening on the whole fundus image. (B) Architecture of ResNet~\cite{ResNet2016} with 50 layers. Each convolutional unit (denoted by a blue block) consists of layers for convolution, batch normalization, and ReLU activation.}
\label{img-SegNet}
\end{figure*}

\subsection{Global Fundus Image Level}

In our DENet, two streams are employed to learn representations on the global fundus image level. The first stream is a standard classification network  by using Residual Network (ResNet)~\cite{ResNet2016}. The ResNet is based on a Convolutional Neural Network and introduces the shortcut connection to handle the vanishing gradient problem  in very deep networks, as shown in Fig.~\ref{img-SegNet} (B). We utilize a ResNet-50 as the backbone model to learn the global representation on the whole fundus image directly, which consists of 5 down-sampling blocks, followed by a global max pooling layer and a fully connected (FC) layer for glaucoma screening. The input image of this stream is resized to  224$\times$224 to enable use of pre-trained model in~\cite{ResNet2016} as initialization for our network.

The second global level stream is the segmentation-guided network, which localizes the optic disc region and produces a detection result based on the disc-segmentation representation. As shown in Fig.~\ref{img-SegNet} (A), the main architecture of the segmentation-guided network is adapted by the U-shape convolutional network (U-Net) in~\cite{Ronneberger2015}, which is an efficient fully convolutional neural network for biomedical image segmentation. Similar to the original U-Net architecture, our method consists of the encoder path (left side) and decoder path (right side). Each encoder path performs convolutional layer with a filter bank to produce a set of encoder feature maps, and the element-wise rectified-linear non-linearity (ReLU) activation function is utilized. The decoder path also utilizes the convolutional layer to output the decoder feature map.  The skip connections transfer the corresponding feature map from encoder path and concatenate them to up-sampled decoder feature maps. Finally, a classifier utilizes $1 \times 1$ convolutional layer with \textit{Sigmoid} activation as the pixel-wise classification to produce the disc probability map. 
Moreover, we extend a new branch from the saddle layer of the U-shape network, where the size scale is the smallest (i.e., 40$\times$40) and the number of channel is the highest (i.e., 512-D). The extended branch acts as an implicit vector with average pooling and flatten layers. Next it connects two fully connected layers to produce a glaucoma classification probability. This pipeline embeds the segmentation-guided representation through the convolutional filters on decoder path of the U-shape network. 
The input image of this stream is resized to  640$\times$640, which guarantees that the image has enough details  to localize disc region accurately.

In our global fundus image level networks, two loss functions are employed. The first one is the binary cross entropy loss function for glaucoma detection layer. The other is the Dice coefficient for assessing disc segmentation~\cite{Crum2006}, which is defined as:
\begin{equation}
	L_{Dice} = 1 -  \dfrac{2 \sum_{i}^{N}  p_{i} g_{i}}{\sum_{i}^{N}  p_{i}^2 + \sum_{i}^{N}  g_{i}^2} ,
\end{equation}
where $N$ is the pixel number, $p_{i} \in [0, 1]$ and $g_{i} \in \{0, 1\} $ denote  predicted probability and binary ground truth label for disc region, respectively. The Dice coefficient loss function can be differentiated yielding the gradient as:
\begin{equation}
\dfrac{\partial L_{Dice} }{\partial p_i }=  \dfrac{ 4 p_i \sum_{i}^{N}  p_i g_i - 2 g_i (\sum_{i}^{N}  p_i^2 + \sum_{i}^{N}  g_i^2)}{(\sum_{i}^{N}  p_i^2 + \sum_{i}^{N}  g_i ^2)^2} .
\end{equation}
These two losses are efficiently integrated into back-propagation via standard stochastic gradient descent (SGD).

Note that we use two phases for training the segmentation-guided model. Firstly, the U-shape segmentation network for disc detection  is trained by pixel-level disc training data with Dice coefficient loss. Then the parameters of CNN layers are frozen and the fully connected layers for the classification task are trained by using glaucoma detection training data.  We use the separate phases to train our segmentation-guided model instead of the multi-task based single stage training, with the following reasons: 1)  Using the disc-segmentation representation on screening could add diversity of the proposed network. 2)  The pixel-level label data for disc segmentation is more expensive than image-level label data for glaucoma detection. The separate stage could employ different training datasets and configuration (e.g, different batch sizes and image numbers).  3) The extracted disc region by network influences the follow-up stream, thus the accuracy of disc detection is more important than classification branch.

\subsection{Optic Disc Region Level}

The second level in our network is based on the local optic disc region, which is cropped based on the previous segmentation-guided network. The local disc region preserves more detailed information with higher resolution and it is benefited to learn a fine representation.  Two local streams are used in our network  to learn representations on the local disc region. The first one is a standard classification network based on ResNet~\cite{ResNet2016} on the original local disc region, as shown in Fig.\ref{img-framework}, while the other stream focuses on the disc polar transformation.

\begin{figure}[!t]
\begin{center}
\includegraphics[width=1\linewidth]{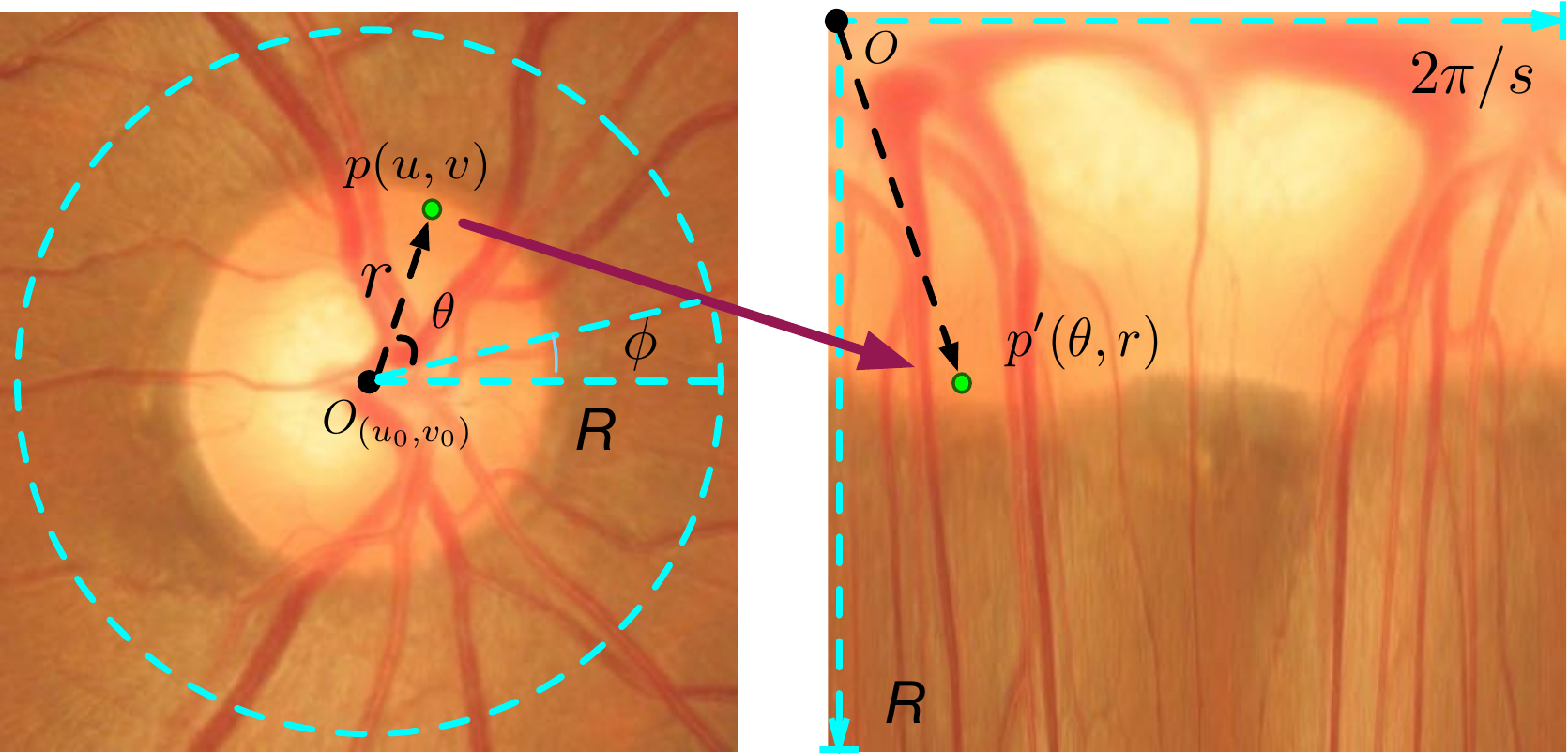}
\caption{Illustration of the mapping from Cartesian coordinate system (Left) to the polar coordinate system (Right) by using the polar transformation. The point $p(u,v)$ in Cartesian coordinate corresponds to the point $p'(\theta, r)$ in polar coordinate.}
\label{img-polar}
\end{center}
\end{figure}

In our method, we apply the pixel-wise polar transformation to transfer the original image to the polar coordinate system. Let $p(u,v)$ denote the point on the original Cartesian plane,  and its corresponding point on polar coordinate system is denoted $p'(\theta, r)$, as shown in Fig.~\ref{img-polar}, where $r$ and $\theta$ are the radius and directional angle of the original point $p$, respectively. Three parameters are utilized to control the polar transformation: the disc center $O(u_o, v_o)$, the polar radius $R$, and the polar angle $\phi$.  The polar transform mapping is calculated as follows:
\begin{equation}
\left\{ {\begin{array}{{l}}
	u = u_o + r \cos (\theta + \phi) , \\
	v = v_o + r \sin (\theta + \phi) ,
	\end{array}}   \right. 
\end{equation}
and the inverse polar transform mapping is as:
\begin{equation}
\left\{ {\begin{array}{{l}}
	r=\sqrt{(u - u_o) ^2 + (v - v_o )^2}  ,  \\
	\theta = \tan ^{-1} (\dfrac{v-v_o}{u-u_o}) - \phi . 
	\end{array}}  \right. \nonumber
\label{Eq_pt}
\end{equation}
The height and width of polar image are set as the polar radius $R$ and discretization $2\pi /s$, where $s$ is the stride. 
The disc polar transformation has the following advantages: 
\begin{enumerate} 
	\item Since the disc and cup are structured as near concentric circles shape. The polar transformation could enlarge the cup region by interpolation. The increased cup region displays more details. As shown in Fig.~\ref{img-polar}, the proportion of cup region is increased and more balanced than that in original fundus image. 
	\item Due to the pixel-wise mapping, the polar transformation is equivariant to the data augmentation on the original fundus image~\cite{Esteves2018}. For example, moving the transformation center $O(u_o, v_o)$ corresponds to the drift  translation in polar coordinates. Using different  polar radius $R$ is the same as augmenting with various scaling factor. And changing the polar angle $\phi$ will shift in  horizontal coordinate in the image. Thus the data augmentation of deep learning can be done by using the polar transformation with various parameters. 
\end{enumerate}

The input disc region of these two local region streams are resized to  224$\times$224 to enable use of pre-trained parameters from ResNet model~\cite{ResNet2016} as initialization. And the binary cross entropy loss function is used for glaucoma detection in the local disc region.

\section{Experiment}
\label{sec_exp}

\subsection{Dataset and Evaluation Criteria}

In our experiments, we use three glaucoma screening datasets, as listed in Table~\ref{Tab_data}. The first one is the ORIGA dataset, which contains 650 fundus images from different eyes, including 168 glaucomatous eyes and 482 normal eyes with manual ground truth of optic disc boundary. The second one is the Singapore Chinese Eye Study  (SCES) dataset, which consists of 1676 images with 46 glaucoma cases. Moreover, a new dataset, called the Singapore Indian Eye Study (SINDI) dataset, is also used. The SINDI dataset is a population-based study conducted also by the Singapore Eye Research Institute, aiming to assess the risk factors of visual impairment in the Singapore Indian community. In this dataset, there is a total of 5783 eye images including 113 glaucomatous eyes and  5670 normal eyes. Only the ORIGA dataset has the manually labelled ground truth of optic disc boundary, thus we use all the 650 images in ORIGA dataset for network training including disc segmentation and glaucoma screening, and employ the SCES and SINDI datasets for testing. The resolution of the fundus image in these three datasets is $3072 \times 2048$, and the size of cropped disc region is $800 \times 800$.

For evaluation, we report the Receiver Operating Characteristic (ROC) curves and the area under ROC curve (AUC). Moreover, we also employ three evaluation criteria to measure performance, including:  Sensitivity (Sen), Specificity (Spe), and Balanced Accuracy (BAcc), which are defined as:
\begin{equation}
\text{Sen} = \dfrac{TP}{TP+FN}, \text{Spe} = \dfrac{TN}{TN+FP}, \text{BAcc} = \dfrac{\text{Sen} + \text{Spe}}{2},  \nonumber 
\end{equation}
where $TP$ and $TN$ denote the number of true positives and true negatives, respectively, and $FP$ and $FN$ denote the number of false positives and false negatives, respectively. Tuning the diagnostic thresholds could obtain a series of criteria scores, and then we use the threshold with highest B-Accuracy score as the final threshold to report the performance.

\begin{table}[!t]
	\centering
	\caption{The datasets used in our  experiments}
	\begin{tabular}{|l|c|c|c|}
		\hline
		Method       &  ORIGA dataset  & SCES dataset & SINDI dataset \\ \hline
		Purpose      &    Training     &   Testing    &    Testing    \\
		Ground Truth & Disc + Glaucoma &   Glaucoma   &   Glaucoma    \\
		Image Num.   &       650       &     1676     &     5783      \\
		Open / Close &    482 / 168    &  1630 / 46   &  5670 / 113   \\ \hline
	\end{tabular} \\
	\label{Tab_data}%
\end{table}%

\begin{table*}[!t]
	\centering
	\caption{Performance comparisons of the different methods on Datasets.}
	\begin{tabular}{|l||c|c|c|c|c|c|c|c|}
		\hline
		                             &        \multicolumn{4}{c|}{SCES Dataset}        &       \multicolumn{4}{c|}{SINDI Dataset}        \\ \hline
		Method                       &  AUC   & B-Accuracy & Sensitivity & Specificity &  AUC   & B-Accuracy & Sensitivity & Specificity \\ \hline
		Airpuff IOP                  & 0.6600 &   0.6452   &   0.3696    &   0.9209    & 0.6233 &   0.5991   &   0.3451    &   0.8531    \\
		Wavelet~\cite{MOOKIAH201273} & 0.6591 &   0.6544   &   0.7174    &   0.5914    & 0.6487 &   0.6262   &   0.6283    &   0.6242    \\
		Gabor~\cite{Acharya2015}     & 0.7594 &   0.7191   &   0.9130    &   0.5252    & 0.7417 &   0.7117   &   0.8053    &   0.6182    \\
		GRI~\cite{Bock2010MIA}       & 0.8051 &   0.7629   &   0.6957    &   0.8301    & 0.7892 &   0.7121   &   0.5752    &   0.8490    \\
		Superpixel~\cite{Cheng2013}  & 0.8265 &   0.7800   &   0.7391    &   0.8209    & 0.7712 &   0.7360   &   0.7257    &   0.7464    \\
		DeepCDR~\cite{Fu2018TMI}     & 0.8998 &   0.8157   &   0.7609    &   0.8706    & 0.7929 &   0.7585   &   0.7522    &   0.7647    \\ \hline
		Image                        & 0.8258 &   0.8011   &   0.7826    &   0.8196    & 0.7806 &   0.7170   &   0.7611    &   0.6728    \\
		Seg                          & 0.8386 &   0.7659   &   0.7391    &   0.7926    & 0.7250 &   0.6656   &   0.6018    &   0.7295    \\
		Disc                         & 0.8592 &   0.8138   &   0.8913    &   0.7362    & 0.7560 &   0.7011   &   0.6018    &   0.8004    \\
		Polar                        & 0.8748 &   0.8294   &   0.8913    &   0.7675    & 0.7663 &   0.7205   &   0.6726    &   0.7684    \\
		Image + Disc                 & 0.8709 &   0.8228   &   0.8696    &   0.7761    & 0.8005 &   0.7481   &   0.7345    &   0.7617    \\
		Image + Seg                  & 0.8832 &   0.8089   &   0.8478    &   0.7699    & 0.7952 &   0.7353   &   0.8319    &   0.6388    \\
		Image + Polar                & 0.8996 &   0.8331   &   0.8913    &   0.7748    & 0.7978 &   0.7437   &   0.7965    &   0.6910    \\
		Disc + Polar                 & 0.8876 &   0.8285   &   0.8913    &   0.7656    & 0.7736 &   0.7204   &   0.6195    &   0.8213    \\
		Disc + Seg                   & 0.8805 &   0.8037   &   0.7609    &   0.8466    & 0.7815 &   0.7173   &   0.8053    &   0.6293    \\
		Seg + Polar                  & 0.8894 &   0.8370   &   0.9348    &   0.7393    & 0.7660 &   0.7121   &   0.8673    &   0.5570    \\
		Image + Disc + Seg           & 0.8965 &   0.8335   &   0.8043    &   0.8626    & 0.8137 &   0.7559   &   0.7788    &   0.7330    \\
		Image + Disc + Polar         & 0.9019 &   0.8479   &   0.9130    &   0.7828    & 0.8060 &   0.7470   &   0.6991    &   0.7949    \\
		Disc + Seg + Polar           & 0.9066 &   0.8520   &   0.9783    &   0.7258    & 0.7912 &   0.7316   &   0.6726    &   0.7907    \\
		Image + Seg + Polar          & 0.9146 &   0.8348   &   0.8696    &   0.8000    & 0.8093 &   0.7564   &   0.7788    &   0.7340    \\ \hline
		Our DENet                    & 0.9183 &   0.8429   &   0.8478    &   0.8380    & 0.8173 &   0.7495   &   0.7876    &   0.7115    \\ \hline
	\end{tabular} \\
	\label{Tab_exp_score}%
\end{table*}%

\begin{figure*}[!t]
	\begin{center}
		\includegraphics[width=0.45\textwidth]{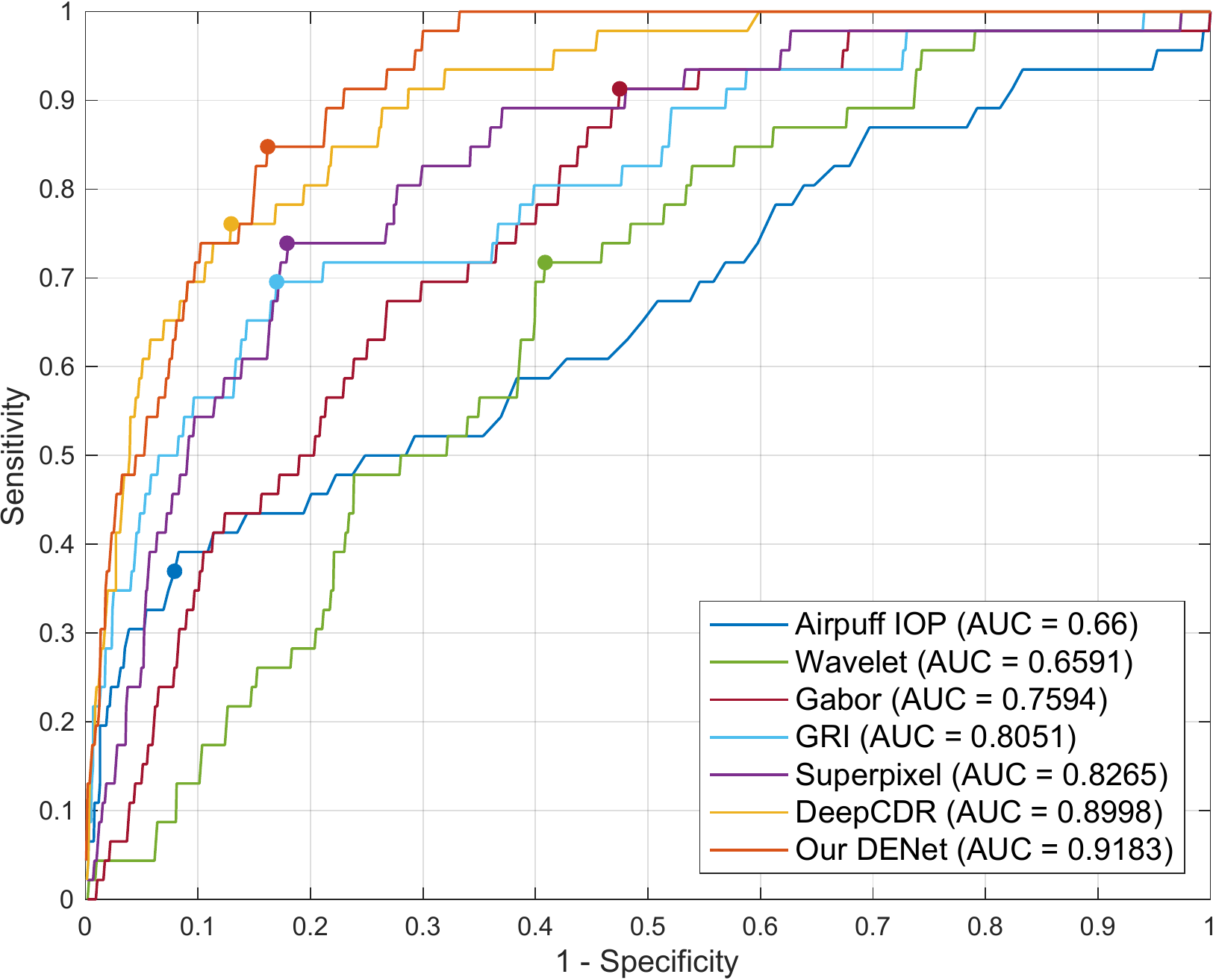}  \;
		\includegraphics[width=0.45\textwidth]{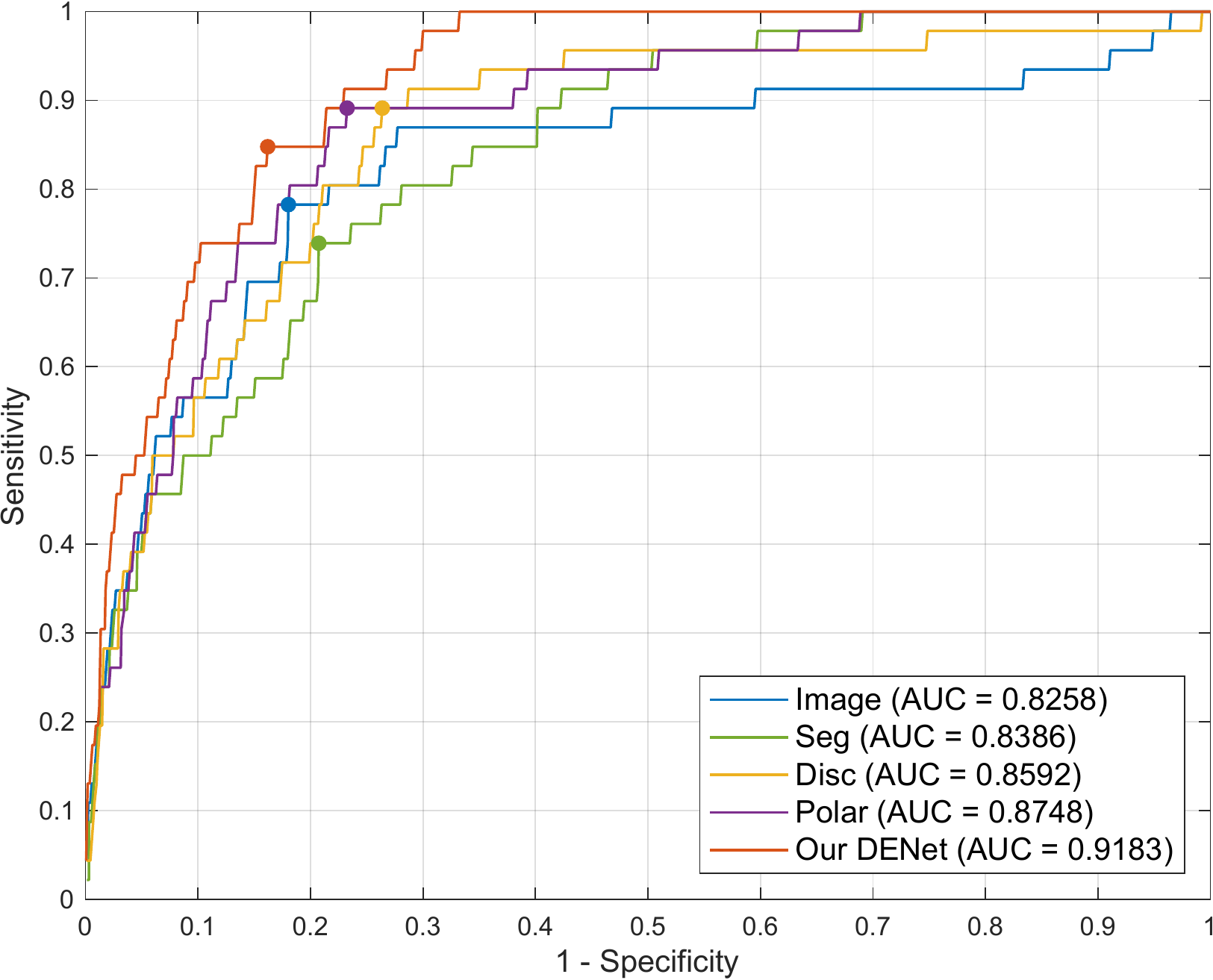} \\ ${}$ \\
		\includegraphics[width=0.45\textwidth]{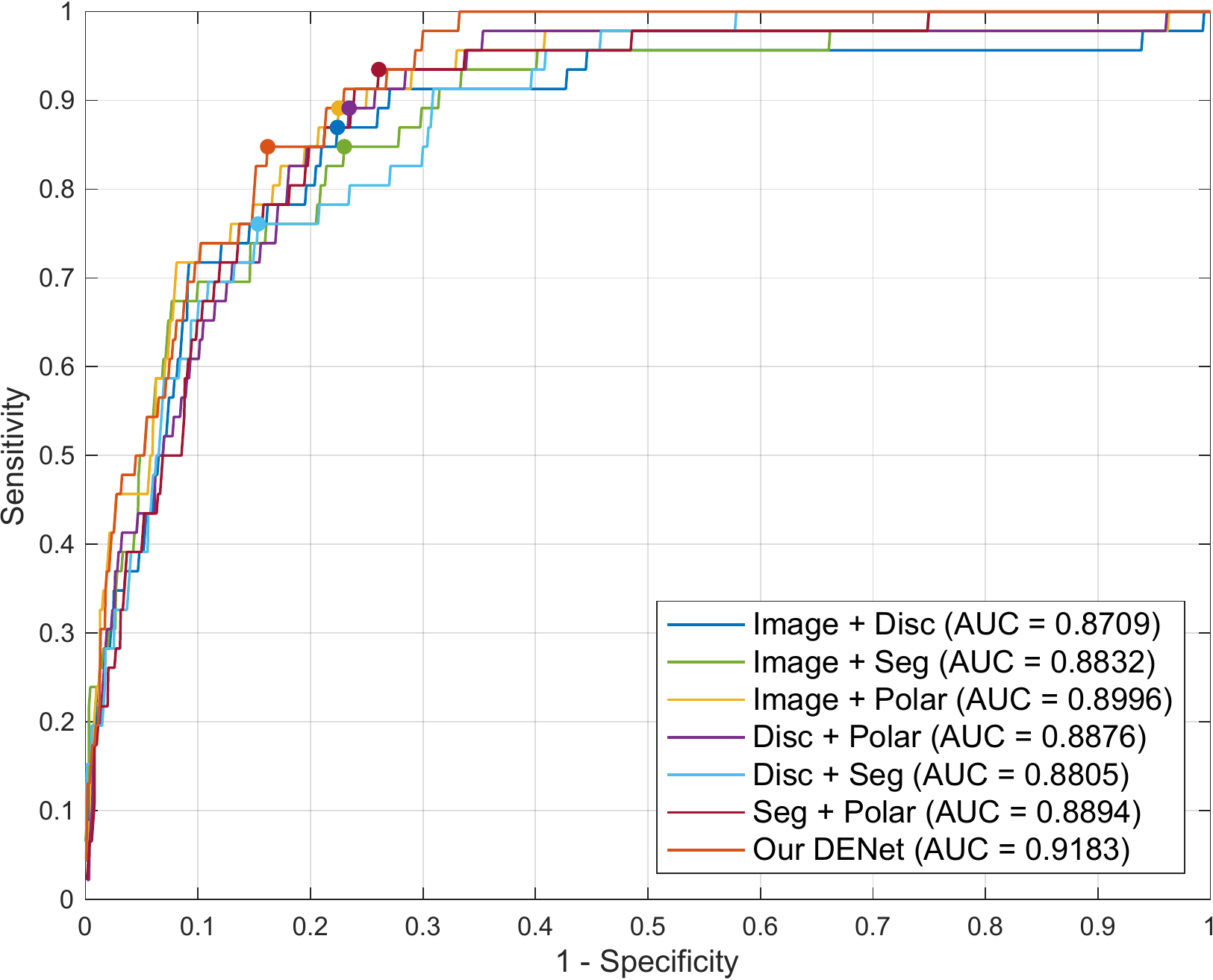}  \;
		\includegraphics[width=0.45\textwidth]{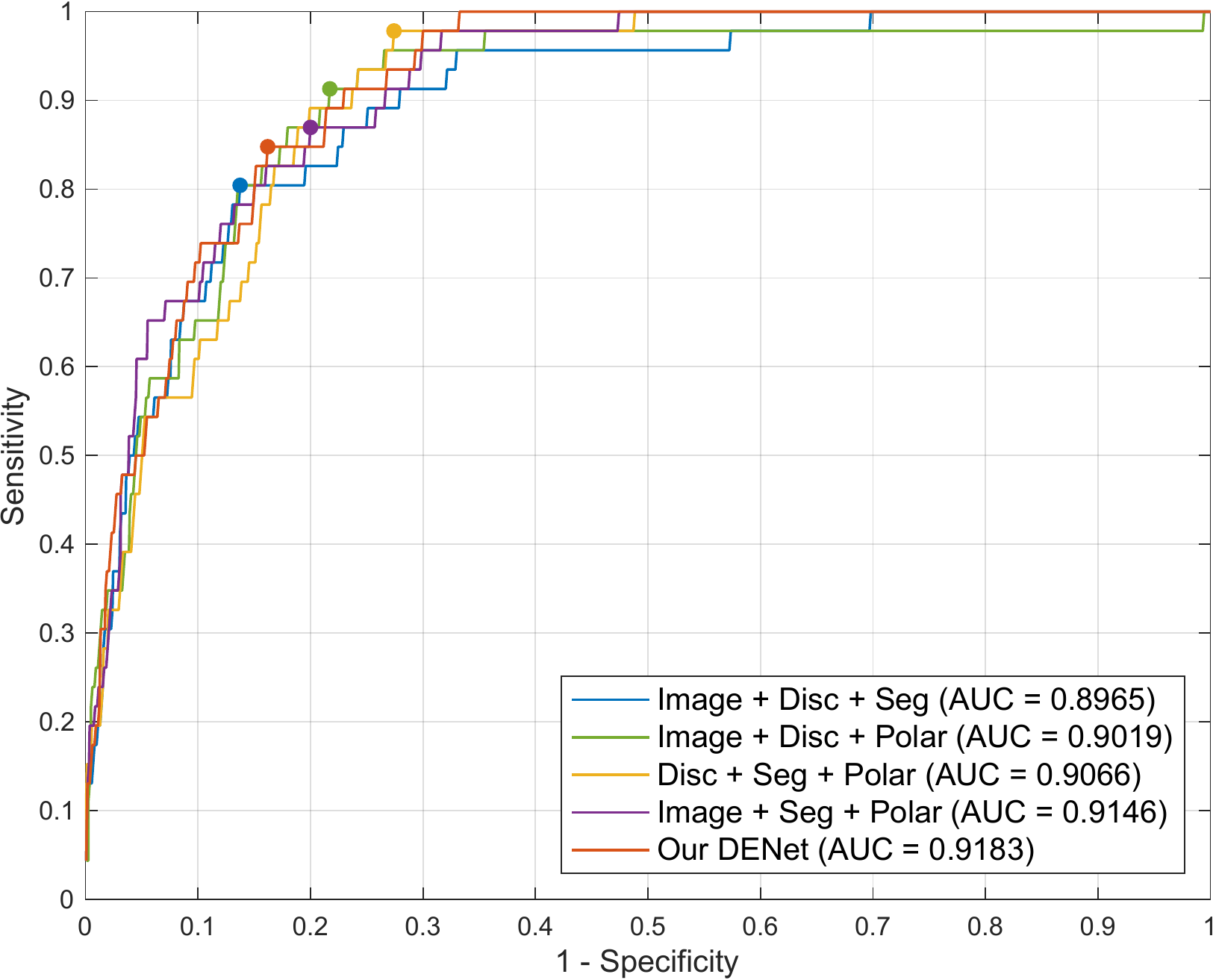}  
		\caption{The ROC curves with AUC scores for glaucoma screening on the SCES dataset, where the  point indicates the highest B-Accuracy score.}
		\label{exp-sces-auc}
	\end{center}
\end{figure*}

\begin{figure*}[!t]
	\begin{center}
		\includegraphics[width=0.45\textwidth]{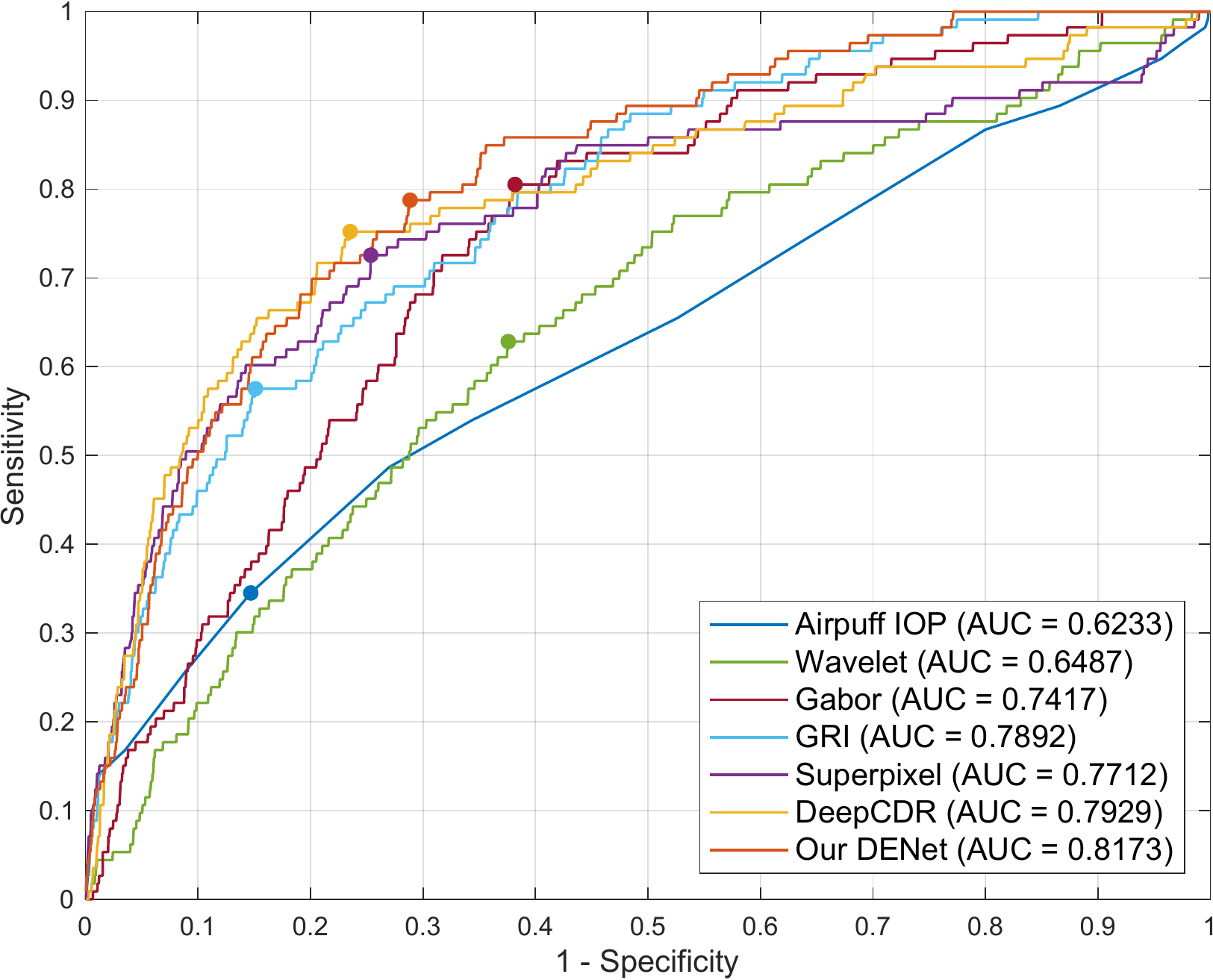} \;
		\includegraphics[width=0.45\textwidth]{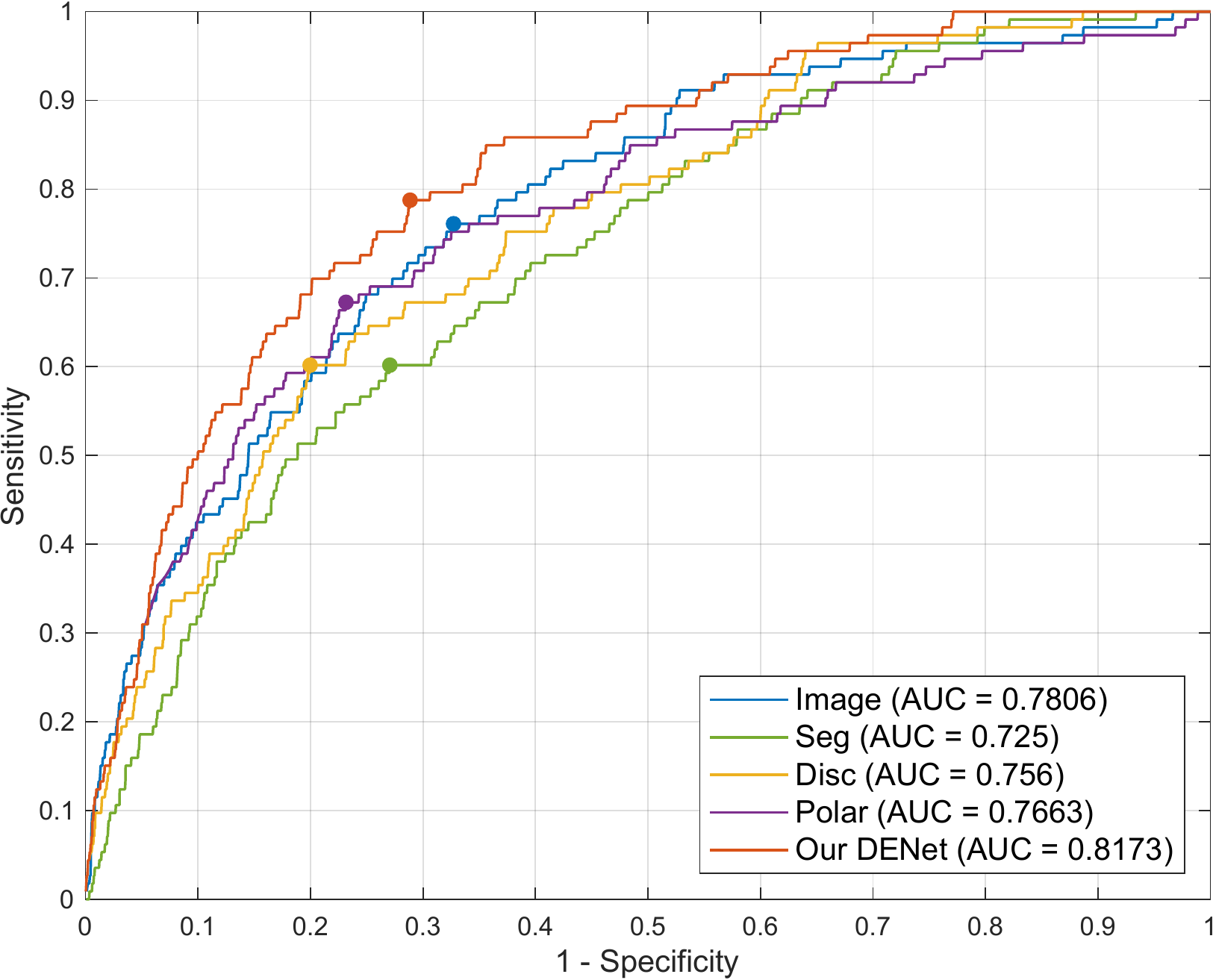} \\  ${}$ \\
		\includegraphics[width=0.45\textwidth]{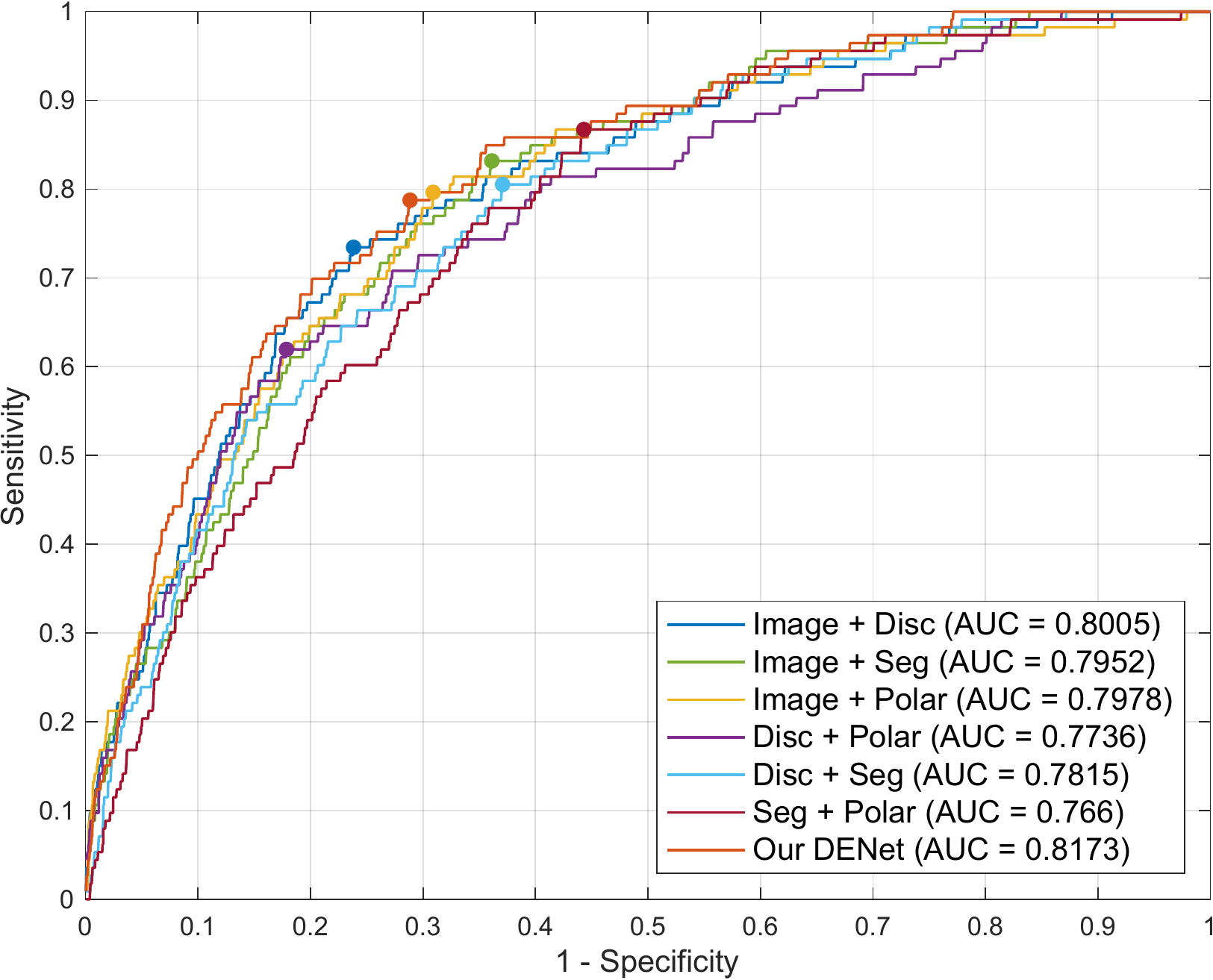} \;
		\includegraphics[width=0.45\textwidth]{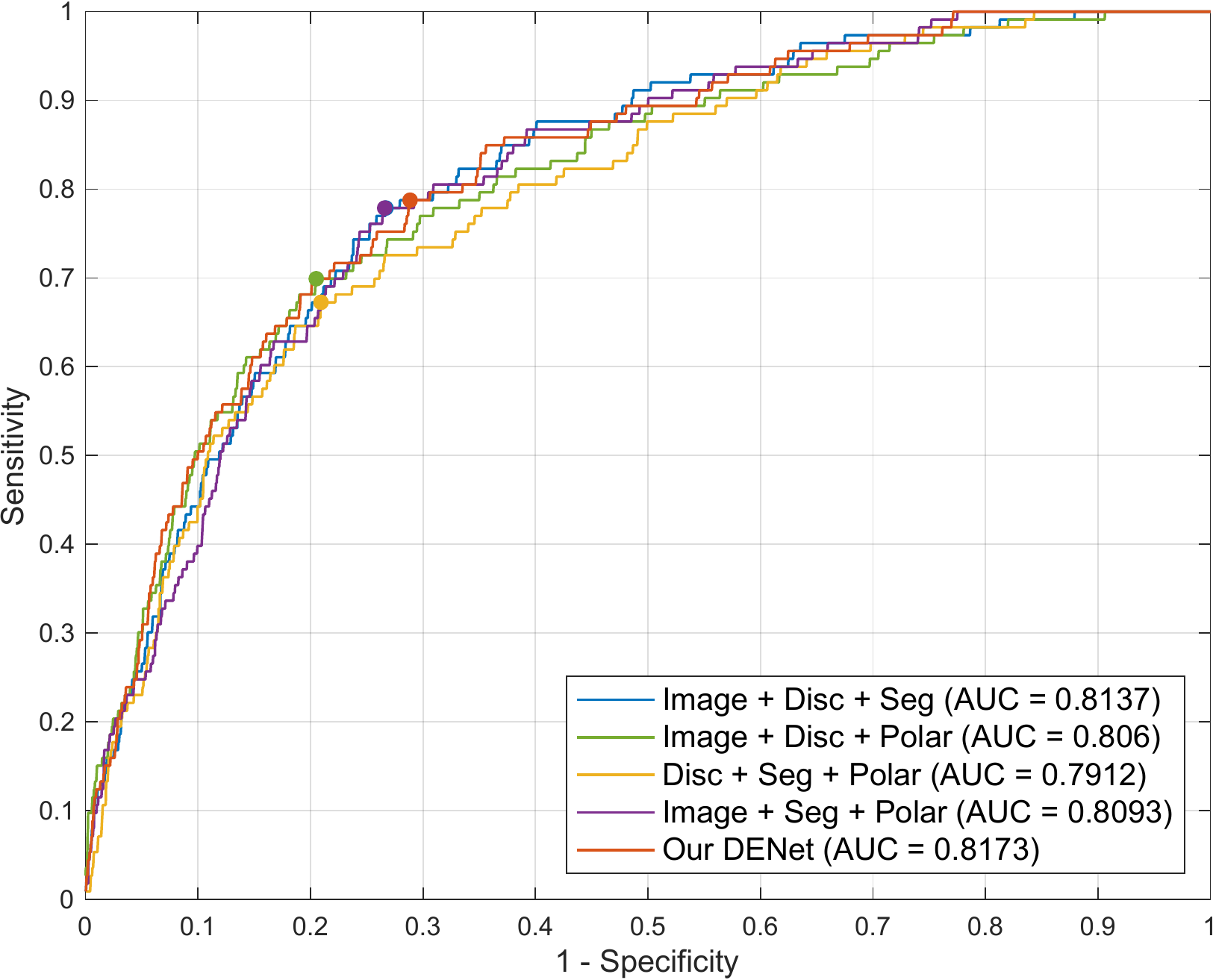}  
		\caption{The ROC curves with AUC scores for glaucoma screening on the SINDI dataset, where the  point indicates the highest B-Accuracy score.}
		\label{exp-sindi-auc}
	\end{center}
\end{figure*}

\subsection{Implementation Details}

Our whole framework is implemented with Python based on Keras with Tensorflow backend. The four streams in our system are trained separately according to different data augmentation strategies, which dues to three reasons: 1) The disc local streams are based on the disc detection result of global image stream. 2) The separate training stage is convenient to add new stream into the ensemble network. 3) The separate training stage could employ different training datasets and configuration for different stream. 
For the global image stream and segmentation-guided network, the data augmentation is done on the training set by random rotations ($0/90/180/270$ degrees) and random flips. For the local disc region scream, the data augmentation is done by random rotations ($0/90/180/270$ degrees), crop drift ($\pm 20$ pixels) and random flips. For the disc polar region, we tune the polar transformation parameters to control the data augmentation, by polar angle  ($\phi = 0/90/180/270$ degrees), polar center drift ($C_{(u_o, v_o)} \pm 20$), and polar radius ($R = 400 \times \{0.8, 1\}$). During training, we employ stochastic gradient descent (SGD) for optimizing the deep model. We use a gradually decreasing learning rate starting from $0.0001$ and a momentum of $0.9$.  Our ResNet-50 stream employs pre-trained parameters based on ImageNet as initialization, and all the layers are fine-tuned during the training.

\subsection{Glaucoma Screening}

We compare our method with five state-of-the-art glaucoma screening baselines: wavelet-based feature method (Wavelet) in~\cite{MOOKIAH201273}, Gabor transformation method (Gabor) in~\cite{Acharya2015}, Glaucoma risk index method (GRI) in~\cite{Bock2010MIA}, superpixel-based classification method (Superpixel)  in~\cite{Cheng2013} and Deep Segmentation method (DeepCDR) in~\cite{Fu2018TMI}. We additional provide the intraocular pressure (IOP) measurement result as the clinical baseline. Moreover, for our algorithm, we show not only the final screening result (Our DENet), but also provide results obtained by each stream, such as the global image stream (Image), segmentation-guided network (Seg), local disc region stream (Disc), and local disc polar stream (Polar). The experiment results are reported in Table~\ref{Tab_exp_score} and Figs.~\ref{exp-sces-auc} and~\ref{exp-sindi-auc}.

From the glaucoma screening results, the IOP performs poorly with  0.66 AUC on the SECS dataset and  0.6233 AUC on the SINDI dataset. The wavelet-based feature method in~\cite{MOOKIAH201273} utilizes the energy property of the  wavelet transformed image, which does not provide enough discriminative capability for glaucoma screening. By contrast the SRI in~\cite{Bock2010MIA} combines multiple image features (e.g., intensity value, FFT coefficient, and B-spline coefficient), and achieves the higher scores than Wavelet in~\cite{MOOKIAH201273}.
The non-deep learning method, Superpixel~\cite{Cheng2013}, produces a competitive performance on the SCES dataset  (0.8265 AUC) and the SINDI dataset (0.7712 AUC), which is better than IOP. The deep learning method, DeepCDR~\cite{Fu2018TMI}, which jointly segments the optic disc and cup, achieves the higher performances on both datasets. Our DENet obtains the best performances on the SCES dataset (0.9183 AUC) and the SINDI dataset (0.8173 AUC), which has about  $2 \%$ improvement on AUC than DeepCDR~\cite{Fu2018TMI}. It can be seen that without extracting  clinical parameters (e.g., CDR), the visual features could be used for glaucoma screening.  One possible reason is that the clinical parameters are based on what information clinicians currently observe, while visual features deeply represent a wider set of image properties, some of which may be unrelated with what clinicians defined explicitly. Hence, visual features gain more latent image  representations, and more discriminative information than clinical parameters. Moreover, our DENet also outperforms other deep learning based methods. For example, the deep learning method in~\cite{Chen2015MICCAI} provides a glaucoma screening system by using CNN feature on the disc region directly, which obtained  0.898 AUC on the SCES dataset. Our method is comparable to that of the deep system~\cite{Chen2015MICCAI}, and is also able to localize the disc region from the whole fundus image.

\subsection{Discussion}

\subsubsection{Stream Analysis}
For each stream of our DENet, the performance  on global image level is lower than that on local disc region. It is reasonable as most clinical symptoms are from region close to the optic disc~\cite{Jonas1999,Garway-Heath352,LESKE200885}. The ResNet-50 has more CNN layers  than our segmentation-guided classification network (50 CNN layers vs 13 CNN layers), which leads the global image stream to get a better performance than segmentation-guided network on the SINDI dataset, and a similar score on the SCES dataset. Another possible reason is that the ResNet loads ImageNet pre-train model as the initialized parameter on training step, while the segmentation-guided network only trains on ORIGA dataset with 650 images without any pre-train model. For local disc region, the polar transformation promotes the representation capability of deep feature, and outperforms original disc region on both datasets. 
Moreover, we also report the different combinations of our streams on the SCES and SINDI datasets. It can be observed that the combinations with three streams have the higher performances than those with two streams. And the combination of all four streams obtains the best scores on both datasets. 

\begin{figure*}[!t]
\centering
\subfigure[SCES dataset]{ \includegraphics[width=0.45\linewidth]{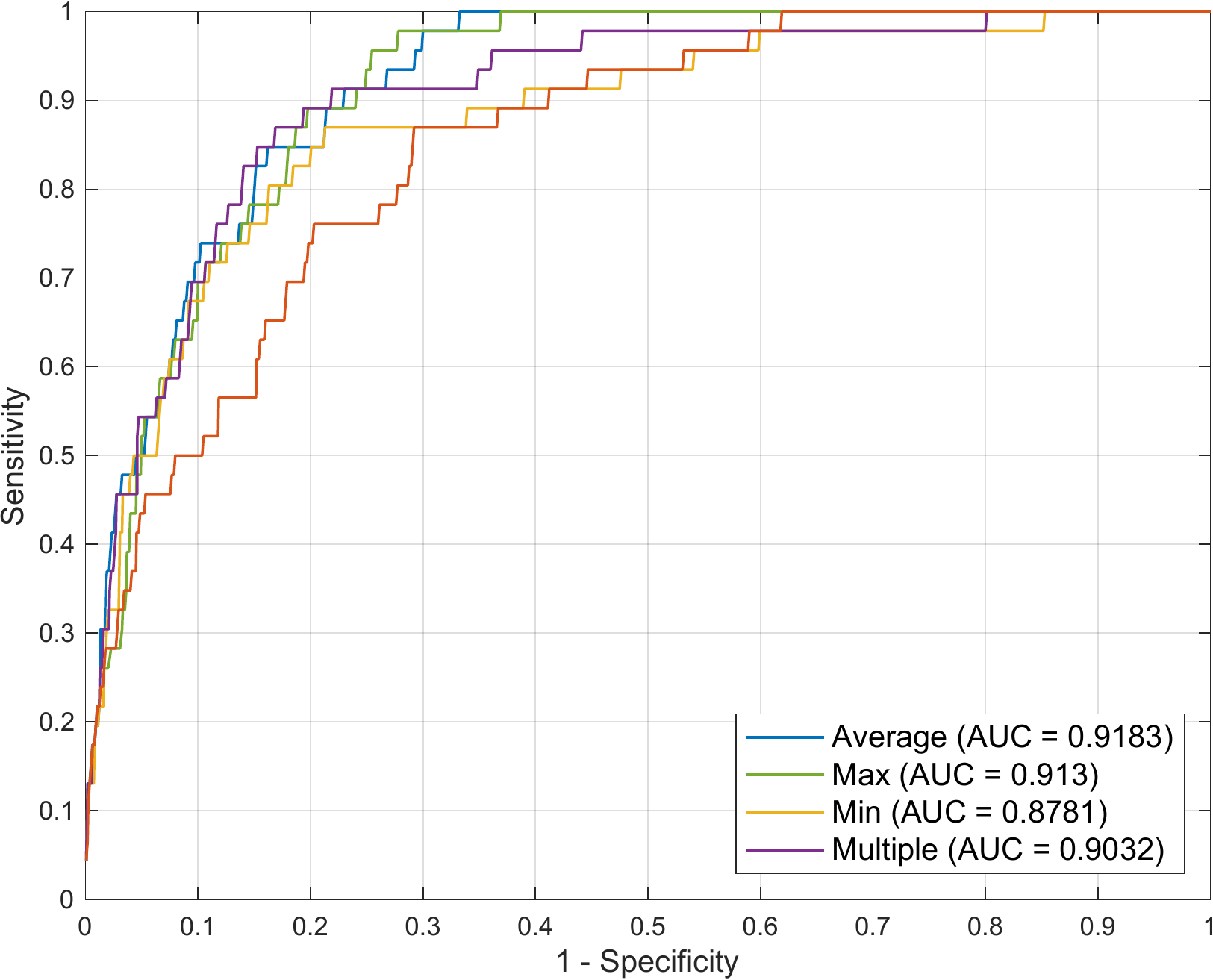}  } \;
\subfigure[SINDI dataset]{ \includegraphics[width=0.45\linewidth]{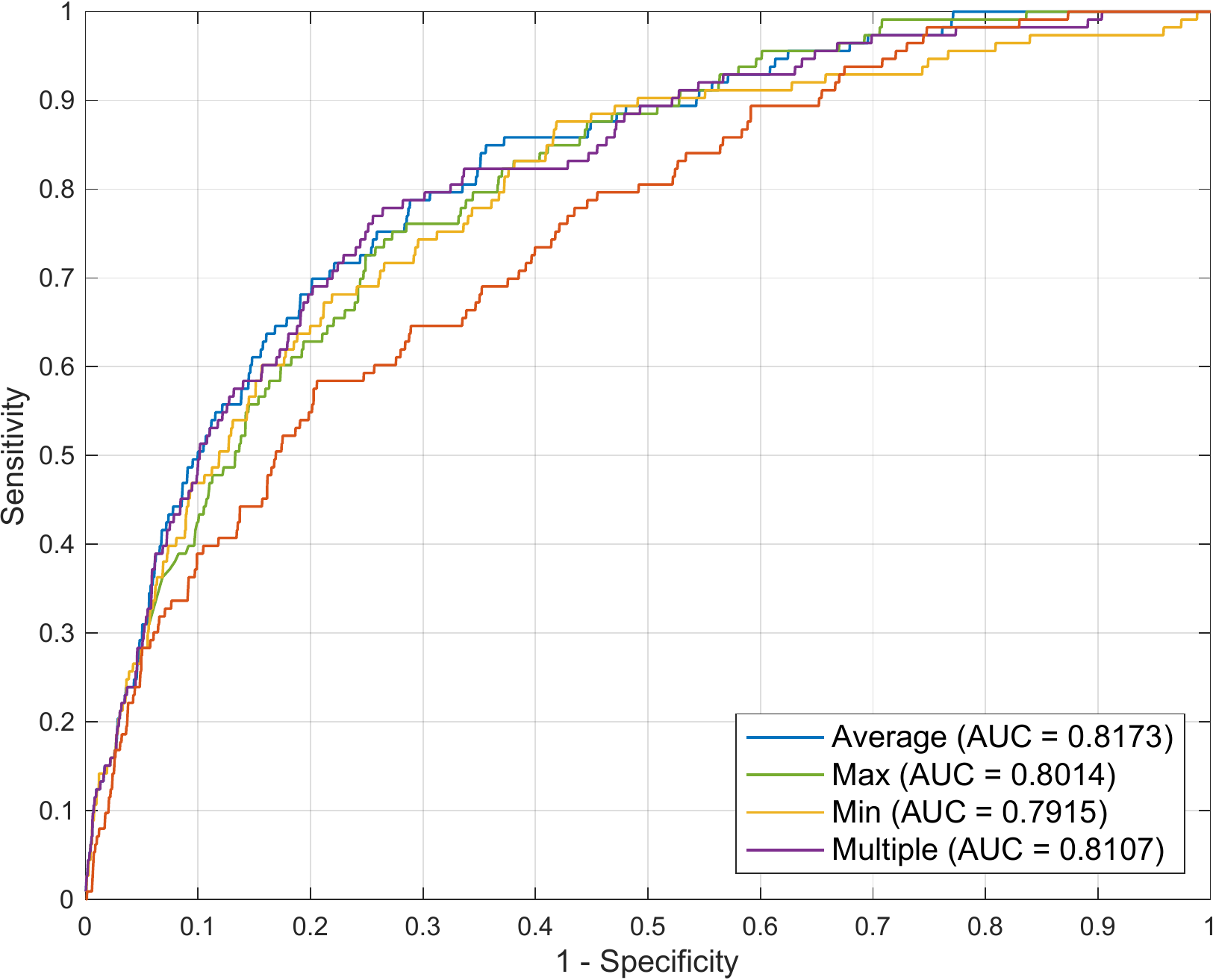} }  
\caption{The performances of the various ensemble operations on the SCES and SINDI datasets.}
\label{exp-fuse}
\end{figure*}

\subsubsection{Ensemble Analysis}
In this experiment, we test different ensemble methods, including average, max, min, and multiple operations. The performances for different ensemble methods on SCES and SINDI datasets are shown in Fig.~\ref{exp-fuse}. Most ensemble operations (e.g., average, max and multiple) result in significant improvements, even the min operation is better than the single stream. This means that the ensemble provides an effective way to improve the screening performance. Moreover,  the  average operation performs best on SCES dataset. Thus,  our DENet uses averaging to ensemble the outputs of four streams.  Note that we do not use the weighted sum to ensemble the streams, as the average operation is a simple and efficient way without any prior information.

\subsubsection{Running Time}
The training phase of each stream takes about 5 hours on a single NVIDIA Titan X GPU (200 iterations). Since we train each stream individually, they could be done parallelly and offline. In testing, our DENet costs only $0.5$ seconds  to generate the final result for one fundus image including disc localization and glaucoma screening, which is faster than the existing measurement-based methods, e.g., superpixel method~\cite{Cheng2013} takes $10$ seconds, DeepCDR~\cite{Fu2018TMI} takes $1$ second. 

\subsubsection{High-Sensitivity Performance}
In real-world, the glaucoma screening system is required to achieve a high sensitivity (not to miss glaucoma case). Thus it would be useful to compare specificities at a certain high sensitivity setting. Table~\ref{Tab_high_sen} shows the performances of the different methods at 0.95 Sensitivity, where our DENet achieves the highest performances on both datasets. 

\begin{table}[!t]
	\centering
	\caption{The performances of different methods at 0.95 sensitivity.}
	\begin{tabular}{|l|c|c|c|c|}
		\hline
		                             & \multicolumn{2}{c|}{SCES Dataset} & \multicolumn{2}{c|}{SINDI Dataset} \\ \hline
		Method                       & B-Accuracy &     Specificity      & B-Accuracy &      Specificity      \\ \hline
		Airpuff IOP                  &   0.4932   &        0.0515        &   0.4953   &        0.0437         \\
		Wavelet~\cite{MOOKIAH201273} &   0.6065   &        0.2564        &   0.5362   &        0.1166         \\
		Gabor~\cite{Acharya2015}     &   0.6415   &        0.3264        &   0.6002   &        0.2446         \\
		GRI~\cite{Bock2010MIA}       &   0.6148   &        0.2730        &   0.6515   &        0.3473         \\
		Superpixel~\cite{Cheng2013}  &   0.6681   &        0.3816        &   0.5018   &        0.0478         \\
		DeepCDR~\cite{Fu2018TMI}     &   0.7700   &        0.5834        &   0.5430   &        0.1303         \\
		Our DENet                    &   0.8316   &        0.7067        &   0.6655   &        0.3753         \\ \hline
	\end{tabular} \\
	\label{Tab_high_sen}%
\end{table}%

\section{Conclusion}
\label{sec_conclusion}

In this paper, we have proposed a novel Disc-aware Ensemble Network (DENet) for automatic  glaucoma screening, which integrates four deep streams on different levels and modules. The multiple levels and modules are beneficial to incorporate the hierarchical representations, while the disc-aware constraint guarantees contextual information from the optic disc region for glaucoma screening. The experiments on two glaucoma datasets (SCES and new collected SINDI datasets) show that our method outperforms  state-of-the-art glaucoma screening algorithms.  The work implementation details are available at \url{http://hzfu.github.io/proj_glaucoma_fundus.html}.

\ifCLASSOPTIONcaptionsoff
\newpage
\fi

{
\bibliographystyle{IEEEtran}
\bibliography{IEEEabrv,Deep_GL}
}

\end{document}